\pdfoutput=1
\PassOptionsToPackage{table}{xcolor}
\documentclass[logo]{sofa_lab_2026}

\graphicspath{{./}}

\usepackage[numbers,sort&compress]{natbib}
\usepackage{bm}
\usepackage{nicefrac}

\usepackage{float}
\usepackage{wrapfig}
\usepackage{tikz}
\usepackage{fontawesome5}
\usepackage{siunitx}
\usepackage{multirow}
\usepackage{adjustbox}
\usepackage{placeins}
\usepackage{xspace}
\usepackage{xparse}
\usepackage{varwidth}
\usepackage[scaled=0.9]{helvet}

\usepackage{amsfonts,amssymb,amsmath}			%
\usepackage{bm}

\usepackage{makecell}
\usepackage{graphics} %
\usepackage{bbold}
\usepackage{graphicx}
\usepackage{tikz}
\usepackage{amsthm}
\usepackage{caption}
\usepackage{subcaption}

\usepackage{bbm}
\usepackage{xspace}
\usepackage{wrapfig}
\usepackage{tabularx}
\usepackage{tablefootnote}
\usepackage{float}
\usepackage{stfloats}
\usepackage{adjustbox}
\usepackage{enumitem}
\usepackage{booktabs} %
\usepackage{mathtools}
\usepackage{multirow}

\newcommand{\ifno}[1]{}

\makeindex
\usepackage{makeidx}
\usepackage{acronym}

\newcommand{\Pc}{\mathcal{P}}

\newcommand{\Vc}{\mathcal{V}}

\newcommand{\Yc}{\mathcal{Y}}

\definecolor{darkgreen}{rgb}{0.0, 0.5, 0.0}
\definecolor{darkred}{rgb}{0.88, 0.09, 0.12}

\newcommand{\be}{\begin{equation}}
\newcommand{\ee}{\end{equation}}
\newcommand{\ba}{\begin{array}}
\newcommand{\ea}{\end{array}}
\newcommand{\bad}{\begin{aligned}}
\newcommand{\ead}{\end{aligned}}

\usetikzlibrary{shapes,arrows}
\usetikzlibrary{arrows,calc}
\tikzset{
    block/.style = {draw, rectangle,
        minimum height=1cm,
        minimum width=2cm},
    input/.style = {coordinate,node distance=1cm},
    output/.style = {coordinate,node distance=3cm},
    arrow/.style={draw, -latex,node distance=2cm},
    pinstyle/.style = {pin edge={latex-, black,node distance=2cm}},
    sum/.style = {coordinate, node distance=1cm}
}

\newcommand{\ours}{{\textsc{Gaze-RL}}}

\usepackage{cleveref}
\crefname{equation}{Eq.}{Eqs.}
\Crefname{equation}{Eq.}{Eqs.}

\definecolor{ink}{HTML}{1A1D23}
\definecolor{ink2}{HTML}{5B616B}
\definecolor{sage}{HTML}{5E7D70}
\definecolor{ivory}{HTML}{F3F5F7}
\definecolor{ivoryline}{HTML}{D4DAE0}
\definecolor{oursbg}{HTML}{DCE5EE}
\definecolor{methodblue}{HTML}{3D5F7E}
\definecolor{pend}{HTML}{8A8F98}
\definecolor{tableInk}{HTML}{263B4A}
\definecolor{tableGroup}{HTML}{F2F5F6}
\definecolor{takeawayline}{HTML}{9CAEBB}
\definecolor{takeawayback}{HTML}{ECEFF2}
\definecolor{takeawaytitleback}{HTML}{DCE2E7}
\colorlet{rule}{BrandRule}
\definecolor{seblue}{HTML}{2F6DB5}

\arrayrulecolor{rule}
\newcommand{\tsz}{\footnotesize}
\newcommand{\tgroup}[2]{\multicolumn{#1}{@{\hspace{2pt}}l}{\cellcolor{tableGroup}\textcolor{tableInk}{\textbf{#2}}}}
\newcommand{\oursrow}{\rowcolor{oursbg}}

\tcbset{
  takeawaysbox/.style={
    title=Contributions,
    width=\linewidth, enhanced, breakable,
    rounded corners, arc=2.8mm, outer arc=2.8mm,
    boxrule=0.7pt,
    top=5pt, bottom=5pt, left=7pt, right=7pt,
    before skip=7pt, after skip=7pt,
    colback=takeawayback, colframe=takeawayline, coltext=black,
    colbacktitle=takeawaytitleback, coltitle=black,
    fonttitle=\bfseries\itshape\small,
    titlerule=0.45pt, toptitle=4pt, bottomtitle=4pt,
    lefttitle=7pt, righttitle=7pt,
  }
}

\newtcbtheorem[auto counter, number within=section]{definition}{Definition}{%
    enhanced,
    theorem style=plain,
    fonttitle=\bfseries\upshape,
    fontupper=\itshape,
    lower separated=false,
    colback=ivory,
    colframe=ivory,
    borderline west={2.2pt}{0pt}{methodblue},
    coltitle=black,
    arc=0mm,
    top=0.4em, bottom=0.4em,
    left=0.6em, right=0.5em,
    before skip=0.8em,
    after skip=0.8em,
    label type=definition,
}{def}
\crefname{definition}{Definition}{Definitions}
\Crefname{definition}{Definition}{Definitions}

\title{Eyes in the Update? Learning Where Rewards Land in Multimodal Reasoning}

\makeatletter
\gdef\@author{%
  \begin{tabular}{@{}c@{}}
  {\Authfont Lixuan He$^{1}$\quad Tianzhe Chu$^{1,3}$\quad Shulin Tian$^{2}$\quad Shikang Zheng$^{1}$\quad Feng Chen$^{1,3}$\quad Yi Ma$^{1,3}$}\\[4pt]
  {\Affilfont $^{1}$The University of Hong Kong\quad $^{2}$Nanyang Technological University\quad $^{3}$TranscEngram}
  \end{tabular}}
\makeatother

\newcommand{\badge}[3]{%
  \mbox{\href{#2}{%
    \IfFileExists{#1}%
      {\raisebox{-0.2\height}{\includegraphics[height=1em]{#1}}\hspace{0.3em}}%
      {}%
    \texttt{#3}%
  }}%
}
\newcommand{\coderepo}{\badge{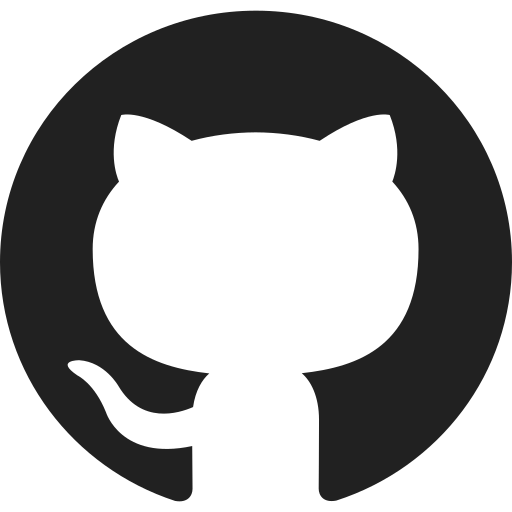}{https://github.com/lixuan27/Gaze-RL-main}{github.com/lixuan27/Gaze-RL-main}}

\teaser{%
  \includegraphics[width=\linewidth]{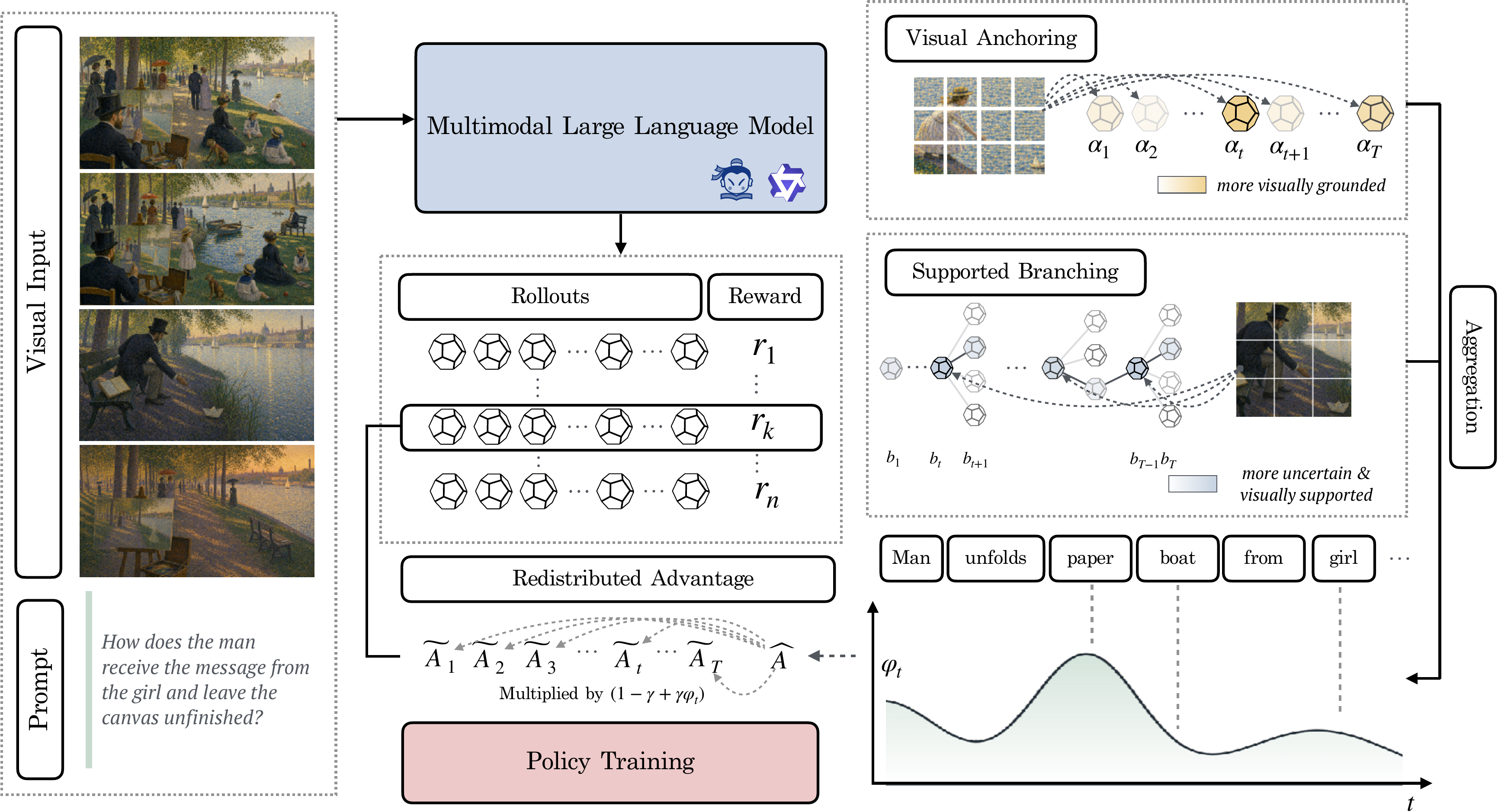}\par
  \captionsetup{skip=4pt}%
  \captionof{figure}{\textbf{Overview of \ours{} for multimodal policy training.} \emph{Visual anchoring} \(\alpha_t\) enters through the support coefficient of \emph{supported branching} \(b_t\) (high predictive entropy with visual support), and \(b_t\) is aggregated into a reward landing field \(\varphi_t\) that redistributes the completion-level advantage \(\widehat A\) into \(\widetilde A_t=\bigl((1-\gamma)+\gamma\varphi_t\bigr)\widehat A\).}%
  \label{fig:teaser}%
}

\fancypagestyle{firststylemail}[firststyle]{%
  \fancyfoot[L]{\footerfont\faEnvelope[regular]\enspace Email: \href{mailto:lixuanhe@stanford.edu}{\texttt{lixuanhe@stanford.edu}}}%
}

\begin{document}
\maketitle

\begin{abstract}
Multimodal post-training has moved from broad instruction tuning toward reinforcement learning for image and video reasoning, where outcome rewards can improve final-answer accuracy. Yet these rewards are assigned to complete generated trajectories, while visual evidence is used only at sparse and uneven positions inside the response. Recent work has begun to make multimodal RL more token-sensitive through grounding signals or external token-level weighting. We ask a more specific question: \textbf{can the allocation of the same trajectory-level reward be inferred from the model's own multimodal computation, rather than from an auxiliary annotator, perception module, or process label?} We propose \ours{}, a representation-driven policy optimization method that redistributes completion-level advantage through a latent reward landing field. The field is computed from internal cross-modal geometry and local predictive uncertainty, assigning larger update weight to positions that are both visually anchored and decision-bearing while remaining mean-one, so the total credit of each rollout is preserved and only its positionwise allocation changes. \ours{} requires no process supervision, auxiliary perception module, or architectural modification, and improves both Qwen2.5-VL and InternVL3.5 across a broad range of image and video benchmarks. Code is available at \coderepo.
\end{abstract}
\thispagestyle{firststylemail}

\clearpage
\setcounter{tocdepth}{2}
\makeatletter\patchcmd{\l@section}{\addvspace{1.0em \@plus\p@}}{\addvspace{0.55em \@plus\p@}}{}{}\makeatother
{\small\setlength{\parskip}{0pt}\tableofcontents}
\clearpage

\section{Introduction}
\label{sec:introduction}

Multimodal large language models (MLLMs) now combine capable image and video perception with post-training for visual reasoning. Stronger backbones, instruction-tuned assistants, and unified image-video models provide the perceptual substrate~\citep{alayrac2022flamingo,li2023blip2,dai2023instructblip,zhu2023minigpt4,liu2023visualinstruction,bai2023qwenvl,llavaonevision2024,tong2024cambrian,qwen25vl2025,internvl3_2025,internvl35_2025}; reinforcement learning and preference optimization increasingly elicit long-form reasoning~\citep{schulman2017proximal,ouyang2022training,rafailov2023direct,ahmadian2024back,shao2024deepseekmath,guo2025deepseek,visualrft2025,visionr1_2025,mmeureka2025,r1onevision2025,r1vl2025,openvlthinker2025,shen2025vlmr1}. Reasoning quality therefore depends on how post-training turns visual representations into evidence-based decisions along a response.

Yet aggregate accuracy does not always imply faithful visual reasoning. Models remain brittle when shortcut cues are suppressed or tasks require expert visual knowledge, diagrams, spatial reasoning, or temporal aggregation~\citep{yu2023mmvet,yue2023mmmu,li2023seedbench,guan2023hallusionbench,mmstar2024,wang2024muirbench,wang2024mathvision,zhang2024mathverse,verify2025,visulogic2025,gong2025space10,videommmu2025,lymperaiou2026patternmatching,adaptmmbench2026,vrb2026,robustvlm2026,frankenstein2026,tian2025morethought}. In videos, decisive evidence may occupy only a few frames or events~\citep{maaz2023videochatgpt,li2023videochat,mangalam2023egoschema,li2023mvbench,videor1_2025,videolmmpost2025,vthinker2025,li2025revisor,videoo3_2026}. A fluent, detailed response can thus contain only a few positions that actually connect visual evidence to the final decision.

Multimodal RL has addressed this problem through curricula, augmentation, cold starts, grounding-aware rewards, and perception-sensitive objectives~\citep{sftorrl2025,noisyrollout2025,grit2025,univgr1_2025,r1sharevl2025,sophiavlr1_2025,rlcoldstart2025,vigorl2025,perceptionr1reward2025,papo2025,vppo2025,vtperceptionr1_2025,nolabels2025,sayo2026,atrl2026}. Related token-sensitive methods shape per-token rewards or reweight policy updates using perception, attention, or process scores~\citep{perceptionr1policy2025,perceptionr1reward2025,grit2025,vigorl2025,rewardmap2025,vppo2025,papo2025}. These methods recognize that generated positions contribute unequally, but often rely on signals external to the policy's reasoning computation. We ask whether the model's internal multimodal states can instead locate token-level credit while preserving the original trajectory-level reward.

We formulate this as \emph{reward landing}: preserve a rollout's total advantage while assigning more update weight to positions that connect visual evidence to consequential decisions. Such positions should be visually anchored in the model's hidden states and decision-bearing, with locally uncertain continuations supported by visually grounded predecessors. \ours{} combines internal cross-modal geometry and predictive uncertainty into a mean-one landing field. It requires no process supervision, auxiliary perception module, or architectural change. Across image and video benchmarks, it improves reasoning-heavy performance and visual evidence sensitivity. The central question is therefore not only how much reward a trajectory receives, but where that reward lands.

\noindent\textbf{Contributions are as follows:}
\begin{tcolorbox}[takeawaysbox]
\begin{itemize}[leftmargin=1em]
    \item We formulate reward landing as allocating fixed trajectory-level credit across response positions that use visual evidence unequally.
    \item We derive a mean-one, token-level landing field from internal cross-modal geometry and predictive uncertainty, without external labels, modules, or architectural changes.
    \item \ours{} improves Qwen2.5-VL and InternVL3.5 on image and video benchmarks and strengthens evidence sensitivity.
\end{itemize}
\end{tcolorbox}

\section{Related Work}
\label{sec:related_work}
\paragraph{Post-training multimodal LLMs (MLLMs) with reinforcement learning.}
RL has progressed from preference alignment to verifier-rewarded reasoning in LLMs~\citep{ouyang2022training,shao2024deepseekmath,guo2025deepseek}. MLLM methods apply outcome-reward algorithms~\citep{shao2024deepseekmath,dapo2025} to visual backbones~\citep{visionr1_2025,visualrft2025,mmeureka2025,r1vl2025,videor1_2025}, often with vision-specific data, perception rewards, or grounding-aware sampling~\citep{perceptionr1policy2025,perceptionr1reward2025,grit2025,vigorl2025,univgr1_2025,r1onevision2025,noisyrollout2025,sftorrl2025}. These approaches improve training but generally broadcast one outcome-level advantage across all response positions. \ours{} instead redistributes that advantage within each rollout while retaining the verifier, data, and architecture.

\paragraph{Improving visual capability in MLLMs.}
Stronger open backbones~\citep{qwen25vl2025,llavaonevision2024,internvl35_2025} and diagnostic benchmarks probe visual indispensability, professional and logical reasoning, and long-form or temporal understanding~\citep{mme2023,mmbench2023,mmvp2024,mmstar2024,singh2019towards,mathew2021docvqa,chartqa2022,mathvista2023,logicvista2024,visulogic2025,verify2025,nextqa2021,videomme2024,tempcompass2024,longvideobench2024,vsibench2024,videommmu2025}. Yet analyses find that RL-trained MLLMs often use visual evidence sparsely along long responses, with aggregate accuracy masking perception-critical failures~\citep{videolmmpost2025,vtperceptionr1_2025,morefinalanswer2025,nolabels2025,vrb2026,adaptmmbench2026,robustvlm2026}. Rather than change the backbone or data, \ours{} targets the update positions implicated by these diagnostics.

\paragraph{Token-level credit assignment in policy updates.}
Multimodal RL can add perception or grounding signals to per-token rewards~\citep{perceptionr1policy2025,perceptionr1reward2025,grit2025,vigorl2025,rewardmap2025} or reweight token-level updates using external perception or attention scores~\citep{vppo2025,papo2025}. Closest to our formulation, PEPO~\citep{pepo2026} reweights token advantages with a softmax over each token's own response--image hidden-state similarity, in which token entropy only modulates the similarity-based exponent through a bounded gate. \ours{} instead weights a position by its predictive entropy times a visual-support coefficient that propagates along a causal kernel over earlier response states, so visual anchoring enters only through the support of the computation behind a position, its credit depends on relations among response states rather than on per-position similarity, and no two weights differ by more than a fixed factor. Its mean-one field changes where rollout credit lands without changing its total.

\section{Preliminaries}
\label{sec:prelim}

\paragraph{Multimodal autoregressive policy.}
The policy $\pi_\theta$ conditions on visual embeddings $\Vc=(v_1,\ldots,v_M)$ and a prompt $\Pc$, then generates $\Yc=(y_1,\ldots,y_T)$. With state $s_t=(\Vc,\Pc,\Yc_{<t})$ and action $a_t=y_t$, its log-likelihood factorizes as \(\log\pi_\theta(\Yc\mid\Vc,\Pc)=\sum_{t=1}^{T}\log\pi_\theta(y_t\mid s_t)\). A verifier assigns an outcome reward $r=R(\Vc,\Pc,\Yc)$. After group normalization, advantage-based methods~\citep{schulman2017proximal,shao2024deepseekmath,dapo2025,vppo2025,ahmadian2024back} yield a completion-level advantage $\widehat A$. Thus, the per-rollout gradient contribution \(\widehat A\nabla_\theta\log\pi_\theta(\Yc\mid\Vc,\Pc)\) applies the same scalar to every token score \(\nabla_\theta\log\pi_\theta(a_t\mid s_t)\).

\paragraph{GRPO.}
GRPO~\citep{shao2024deepseekmath} estimates advantages without a learned value function. For each input $(\Vc,\Pc)$, it samples $G$ responses $\{\Yc_j\}_{j=1}^{G}$ from $\pi_{\theta_{\mathrm{old}}}$, collects rewards $r_j=R(\Vc,\Pc,\Yc_j)$, and computes
\begin{equation}
    \widehat A_j
    =\frac{r_j-\operatorname{mean}_{g\in[G]}r_g}
           {\operatorname{std}_{g\in[G]}r_g+\delta},
    \label{eq:group_adv}
\end{equation}
where $\delta>0$ stabilizes normalization. \ours{} uses GRPO, although landing-field reweighting applies to any method with a similar per-rollout advantage. For token $y_t$, the probability ratio is \(\rho_t(\theta)=\pi_\theta(y_t\mid s_t)/\pi_{\theta_{\mathrm{old}}}(y_t\mid s_t)\), and the reference-policy penalty is \(\mathrm{KL}_t(\theta)=D_{\mathrm{KL}}(\pi_\theta(\cdot\mid s_t)\Vert\pi_{\mathrm{ref}}(\cdot\mid s_t))\), where $\pi_{\mathrm{ref}}$ is fixed. The clipped surrogate is \(\psi_\epsilon(\rho,A)=\min\{\rho A,\operatorname{clip}(\rho,1-\epsilon,1+\epsilon)A\}\), with $\epsilon\in(0,1)$. The GRPO objective is
\begin{equation}
    \mathcal{J}_{GRPO}(\theta)
    =\mathbb{E}_{\mathrm{rollout}}\!\left[
        \frac{1}{T}\sum_{t=1}^{T}
        \bigl(\psi_\epsilon(\rho_t(\theta),\widehat A)
        -\beta\,\mathrm{KL}_t(\theta)\bigr)
    \right],
    \label{eq:prelim_grpo}
\end{equation}
with $\beta\ge0$ controlling reference-policy regularization. The method below changes how $\widehat A$ is distributed across token-level update terms.

\section{Method}
\label{sec:method}

GRPO (\Cref{eq:prelim_grpo}) weights every response position by the same completion-level advantage \(\widehat A\). Yet visual evidence and consequential reasoning decisions are not used uniformly along a rollout: some positions ground their next-token choice in \(\Vc\), others drive a branching decision, and many do neither. \ours{} therefore replaces this uniform broadcast with a token-level redistribution --- a reward landing field that converts \(\widehat A\) into a landed advantage \(\widetilde A_t\) before policy improvement.

\begin{definition}{Reward Landing Field}{landing_field}
    Given a rollout \(\Yc\) of length \(T\) with completion-level advantage \(\widehat A\), a \emph{reward landing field} is a function \(\Phi:[T]\to\mathbb R_{\ge 0}\) with
    \begin{equation}
        \Phi(t)\ge0,
        \qquad
        \frac{1}{T}\sum_{t=1}^{T}\Phi(t)=1,
        \qquad
        \widetilde A_t=\Phi(t)\,\widehat A.
        \label{eq:landing_field}
    \end{equation}
\end{definition}

The mean-one constraint preserves the rollout's total scalar advantage, \(\sum_t \widetilde A_t=T\widehat A\), but does not imply identical clipped policy gradients; the uniform field \(\Phi\equiv 1\) recovers GRPO. The remainder of this section constructs a rollout-dependent \(\Phi\) from two signals --- visual anchoring (\Cref{subsec:method_visual_anchoring}) and visually supported branching (\Cref{subsec:method_supported_branching}) --- and plugs it into the policy update through a landed objective (\Cref{subsec:method_objective}).
\subsection{Visual anchoring}
\label{subsec:method_visual_anchoring}

Visual anchoring provides the first signal for position-wise weighting.
In standard advantage-based policy training, the same \(\widehat A\) is applied to every token-level update term as in \Cref{eq:prelim_grpo}, even though visual evidence is usually used unevenly across the generated response.
We therefore assign each response position \(t\) a rollout-local anchoring score that measures the similarity between its pre-action hidden state and the hidden states associated with the visual embeddings \((v_1,\dots,v_M)\).
This score is not an additional reward and does not require process supervision; it is used only to distribute the already computed advantage \(\widehat A\) across response tokens.

Let \(\mathcal L\) denote the internal transformer layers of the multimodal policy \(\pi_\theta\) used for multimodal reasoning.
For layer \(\ell\in\mathcal L\), let \(h^{\mathrm{txt}}_{\ell,t}\in\mathbb R^d\) be the response-side state aligned to \(s_t\) before predicting \(y_t\), and let \(h^{\mathrm{vis}}_{\ell,m}\in\mathbb R^d\) be the hidden state associated with visual embedding \(v_m\).
For each response position, we score how close this state is to the visual hidden states, averaged over the selected layers:
\begin{equation}
    \alpha_t
    =
    \frac{1}{|\mathcal L|}
    \sum_{\ell\in\mathcal L}
    \left\langle
        \frac{h^{\mathrm{txt}}_{\ell,t}}{\|h^{\mathrm{txt}}_{\ell,t}\|_2},
        \frac{1}{M}\sum_{m=1}^{M}
        \frac{h^{\mathrm{vis}}_{\ell,m}}{\|h^{\mathrm{vis}}_{\ell,m}\|_2}
    \right\rangle,
    \qquad
    \widehat\alpha_t
    =
    \operatorname{Norm}_{1:T}(\alpha_t).
    \label{eq:visual_anchor_score}
\end{equation}
Here \(\operatorname{Norm}_{1:T}\) denotes rollout-wise min-max scaling over response-token positions.
Thus, \(\widehat\alpha_t\in[0,1]\) is the visual anchoring signal passed to the support coefficient of \Cref{eq:branch_coeff}, with larger values indicating stronger visual anchoring relative to the rest of the same rollout.
For Qwen2.5-VL-7B we use layers 14, 18, 22, and 26 and fix them across tasks.

\subsection{Visually supported branching}
\label{subsec:method_supported_branching}

Visually supported branching provides the second signal for position-wise weighting.
Visual anchoring marks positions that draw on the visual context, but it does not distinguish those that drive consequential reasoning decisions: an anchored position with a near-deterministic next-token distribution is unlikely to be where policy improvement should land.
We therefore start from the rollout-time predictive entropy \(H_t=\operatorname{Ent}(\pi_{\theta_{\mathrm{old}}}(\cdot\mid s_t))\) (\Cref{eq:local_entropy}), which is high where the generation decision is locally uncertain, and keep only the uncertainty that is supported by visual evidence through a visual-support coefficient \(\zeta_t\in[0,1]\).
Like visual anchoring, this score is not an additional reward and requires no process supervision.

The coefficient asks whether position \(t\) is visually anchored itself or draws on earlier response states that are.
Each position is summarized by its layer-averaged pre-action state, and each predecessor \(\tau<t\) is weighted by a causal kernel that combines state similarity with a log-distance decay:
\begin{equation}
    z_t=\frac{\bar h_t}{\|\bar h_t\|_2+\varepsilon_{\mathrm{sup}}},
    \quad
    \bar h_t=\frac{1}{|\mathcal L|}\sum_{\ell\in\mathcal L}h^{\mathrm{txt}}_{\ell,t},
    \qquad
    K_{\tau t}
    =
    \frac{\exp\!\bigl(\langle z_\tau,z_t\rangle/\kappa-\lambda_{\mathrm{lag}}\log(1+t-\tau)\bigr)}
         {\sum_{j<t}\exp\!\bigl(\langle z_j,z_t\rangle/\kappa-\lambda_{\mathrm{lag}}\log(1+t-j)\bigr)},
    \label{eq:branch_kernel}
\end{equation}
where \(\kappa>0\) is a temperature and \(\lambda_{\mathrm{lag}}\ge0\) controls distance decay.
The support is defined recursively along the kernel,
\begin{equation}
    \zeta_t=(1-\chi)\,\widehat\alpha_t+\chi\sum_{\tau<t}K_{\tau t}\,\zeta_\tau,
    \label{eq:branch_coeff}
\end{equation}
with propagation share \(\chi\in[0,1)\).
Because \(K\) is strictly causal and each of its rows sums to at most one, \(\zeta\) is obtained by one forward substitution and stays in \([0,1]\); it is large when a position is anchored itself or draws on supported predecessors, and \(\chi=0\) reduces it to the position's own anchoring.
The visually supported branching score is
\begin{equation}
    b_t = H_t\,\zeta_t,
    \label{eq:branch_score}
\end{equation}
which is passed to the landing-field aggregator and is large only when an uncertain position is visually anchored or supported by visually anchored predecessors.
Because the field is computed from the response states themselves, the credit of a position depends on its relations to earlier response states, not only on its own similarity and entropy.

\subsection{Aggregation and landed policy optimization}
\label{subsec:method_objective}

The reward landing field aggregates the supported branching score \(b_t\) of \Cref{eq:branch_score} into a single mean-one weighting over response positions.
We define the log-evidence \(e_t=\log b_t\) and its rollout-centered version \(\tilde e_t=e_t-\frac{1}{T}\sum_{\tau}e_\tau\), and construct the landing field as
\begin{equation}
    \varphi_t
    =
    T\,
    \frac{
        \exp\!\bigl(\eta\tanh\tilde e_t\bigr)
    }{
        \sum_{\tau=1}^{T}
        \exp\!\bigl(\eta\tanh\tilde e_\tau\bigr)
    },
    \label{eq:landing_field_softmax}
\end{equation}
where \(\eta\ge0\) is the modulation strength and the prefactor \(T\) enforces the mean-one constraint of \Cref{eq:landing_field}; scores are floored at \(10^{-12}\) before the logarithm.
The bounded exponent \(\eta\tanh\tilde e_t\in[-\eta,\eta]\) raises the weight of positions whose supported branching is above the rollout's geometric mean and lowers it for those below, so no two weights differ by more than a factor \(e^{2\eta}\); setting \(\eta=0\) recovers the uniform field.
Visual anchoring enters through \(\zeta_t\), so a position is weighted by its uncertainty and by the visual support of the computation behind it.

Both the uniform field \(\Phi\equiv 1\) and the constructed field \(\varphi_t\) of \Cref{eq:landing_field_softmax} satisfy the mean-one constraint of \Cref{eq:landing_field}, so any convex combination is also a valid landing field. We deploy the one-parameter blend
\begin{equation}
    \Phi_\gamma(t)=(1-\gamma)+\gamma\,\varphi_t,
    \qquad \gamma\in[0,1],
    \label{eq:landing_blend}
\end{equation}
in which \(\gamma\) is the \emph{landing strength}: \(\gamma=0\) collapses \(\Phi_\gamma\) to the uniform field and recovers GRPO, while \(\gamma=1\) uses the full constructed field. The total rollout credit \(\sum_t\widetilde A_t=T\widehat A\) is preserved for every \(\gamma\).
Our defaults are \(\gamma=0.70\), \(\eta=0.25\), \(\kappa=0.50\), \(\lambda_{\mathrm{lag}}=0.10\), and \(\chi=0.25\), so that \(\Phi_\gamma(t)\ge1-\gamma(1-e^{-2\eta})\approx0.72\) at every position. The full causal kernel costs \(O(T^2d)\) during training.

\paragraph{Limiting behavior of the support coefficient.}
Since \Cref{eq:landing_field_softmax} is invariant to a common rescaling of \(b_t\), \(\zeta_t\) acts only through its variation across positions: it re-ranks uncertain positions by their visual support rather than switching entropy off, and any constant positive \(\zeta_t\) reduces the field to one of entropy alone.
\Cref{app:limiting_behavior} proves this and derives the bounds of the field used above. The variant \emph{entropy without visual support} in \Cref{fig:ablation_and_sweep} trains the constant-\(\zeta_t\) limit directly.

\ours{} then feeds the landed advantage \(\widetilde A_t=\Phi_\gamma(t)\,\widehat A\) into the clipped policy surrogate \(\psi_\epsilon\) defined in \Cref{sec:prelim}, in place of the completion-level \(\widehat A\) used by GRPO in \Cref{eq:prelim_grpo}, while leaving the verifier, group-relative advantage, clipping radius \(\epsilon\), and reference-policy penalty \(\beta\,\mathrm{KL}_t\) untouched,
\begin{equation}
    \mathcal J_{\ours}(\theta)
    =
    \mathbb E_{\mathrm{rollout}}\!\left[
        \frac{1}{T}\sum_{t=1}^{T}\!\Bigl(
            \psi_\epsilon\bigl(\rho_t(\theta),\widetilde A_t\bigr)
            -\beta\,\mathrm{KL}_t(\theta)
        \Bigr)
    \right],
    \label{eq:ours_objective}
\end{equation}
with \(\rho_t(\theta)\) and \(\mathrm{KL}_t(\theta)\) defined in \Cref{sec:prelim}.
The only modification relative to GRPO is the landing field \(\Phi_\gamma\), which redistributes the completion-level advantage across token-level update terms while preserving the total scalar credit assigned to each rollout. We compute it from rollout-time old-policy states and detach it before optimization; the sign of \(\widehat A\) still determines reinforcement or penalty (\Cref{app:method_details}).

\section{Experiments}
\label{sec:experiments}

\begin{table*}[t]
\centering
\tsz
\setlength{\tabcolsep}{5pt}
\renewcommand{\arraystretch}{1.06}
\caption{\textbf{General image benchmarks.} MME uses the original summed score; all other entries are accuracies or official benchmark scores. Higher is better; \textbf{bold} marks the best result within each backbone.}
\label{tab:image_general}
\begin{adjustbox}{max width=\textwidth}
\begin{tabular}{@{}lccccccc@{}}
\toprule
\textbf{Model and method} & \textbf{MME} & \textbf{MMVP} & \textbf{MMBench} & \textbf{MMStar} & \textbf{CV-Bench} & \textbf{POPE} & \textbf{RefCOCO} \\
\midrule
\tgroup{8}{Qwen2.5-VL-3B} \\
Base model & 2151.2 & 65.7 & 76.4 & 54.1 & 70.2 & 83.2 & 83.9 \\
+ GRPO & 2148.6 & 67.3 & 77.0 & 54.5 & 71.9 & 84.1 & 85.2 \\
+ DAPO & 2152.2 & 68.0 & 77.4 & 54.9 & \textbf{73.2} & \textbf{85.3} & 85.9 \\
+ VPPO & 2149.0 & 66.3 & 77.6 & 53.6 & 71.5 & 83.5 & 85.1 \\
\oursrow + \ours{} (ours) & \textbf{2165.0} & \textbf{68.3} & \textbf{78.3} & \textbf{56.1} & 72.6 & 85.1 & \textbf{86.8} \\
\tgroup{8}{Qwen2.5-VL-7B} \\
Base model & 2309.5 & 76.3 & 82.1 & 62.6 & 73.6 & 85.9 & 85.1 \\
+ GRPO & 2309.7 & 76.7 & 82.7 & 63.2 & 74.5 & 87.0 & 86.7 \\
+ DAPO & 2312.7 & 77.0 & 83.4 & 64.1 & 75.1 & 87.4 & 87.3 \\
+ VPPO & 2313.2 & 76.3 & 83.3 & 63.8 & 73.7 & 86.5 & 87.2 \\
\oursrow + \ours{} (ours) & \textbf{2344.1} & \textbf{78.7} & \textbf{84.0} & \textbf{64.4} & \textbf{75.9} & \textbf{88.0} & \textbf{87.9} \\
\tgroup{8}{InternVL3.5-8B} \\
Base model & 2364.6 & 76.0 & 79.5 & 67.6 & 77.0 & 87.7 & 88.2 \\
+ GRPO & 2360.3 & 77.3 & 81.1 & 68.9 & 78.4 & 88.1 & 88.8 \\
+ DAPO & 2367.3 & 78.0 & 81.9 & \textbf{70.5} & 79.0 & 88.5 & 89.2 \\
+ VPPO & 2362.7 & 76.0 & 81.6 & 69.8 & 77.2 & 87.8 & 88.7 \\
\oursrow + \ours{} (ours) & \textbf{2371.7} & \textbf{79.3} & \textbf{83.0} & 70.1 & \textbf{80.8} & \textbf{89.0} & \textbf{90.1} \\
\bottomrule
\end{tabular}
\end{adjustbox}
\end{table*}

\begin{table*}[t]
\centering
\tsz
\setlength{\tabcolsep}{5pt}
\renewcommand{\arraystretch}{1.06}
\caption{\textbf{OCR-heavy and reasoning-heavy image benchmarks.} All entries follow the official evaluation protocol of each benchmark. Higher is better; \textbf{bold} marks the best result within each backbone.}
\label{tab:image_reasoning}
\begin{adjustbox}{max width=\textwidth}
\begin{tabular}{@{}lccccccc@{}}
\toprule
\textbf{Model and method} & \textbf{Geometry3K} & \textbf{TextVQA} & \textbf{DocVQA} & \textbf{AI2D} & \textbf{ChartQA} & \textbf{MathVista} & \textbf{LogicVista} \\
\midrule
\tgroup{8}{Qwen2.5-VL-3B} \\
Base model & 15.3 & 78.3 & 93.5 & 81.1 & 80.8 & 60.9 & 36.6 \\
+ GRPO & 25.0 & 78.8 & 93.9 & 81.7 & 81.4 & 62.4 & 37.5 \\
+ DAPO & 26.8 & 79.4 & 94.1 & 82.1 & 81.8 & 62.9 & 38.7 \\
+ VPPO & 26.4 & 79.2 & 93.8 & 81.9 & 81.1 & 62.0 & 37.3 \\
\oursrow + \ours{} (ours) & \textbf{29.1} & \textbf{79.9} & \textbf{94.2} & \textbf{82.8} & \textbf{83.0} & \textbf{64.1} & \textbf{39.1} \\
\tgroup{8}{Qwen2.5-VL-7B} \\
Base model & 36.8 & 84.0 & 95.1 & 83.2 & 83.1 & 67.2 & 42.9 \\
+ GRPO & 39.6 & 84.5 & 94.8 & 84.2 & 84.0 & 68.6 & 44.7 \\
+ DAPO & 41.9 & 84.9 & 95.3 & \textbf{85.3} & \textbf{84.7} & 69.4 & 45.8 \\
+ VPPO & 44.0 & 84.7 & 94.4 & 84.4 & \textbf{84.7} & 66.7 & \textbf{46.1} \\
\oursrow + \ours{} (ours) & \textbf{47.9} & \textbf{85.6} & \textbf{95.8} & 85.0 & 84.5 & \textbf{70.4} & \textbf{46.1} \\
\tgroup{8}{InternVL3.5-8B} \\
Base model & 40.3 & 77.6 & 91.8 & 83.5 & 82.7 & 76.5 & 57.3 \\
+ GRPO & 43.2 & 78.3 & 92.1 & 84.1 & 84.4 & 77.3 & 60.8 \\
+ DAPO & 45.0 & 78.6 & 92.3 & 85.0 & 84.8 & 78.7 & 61.6 \\
+ VPPO & 44.8 & 78.0 & 91.9 & 84.6 & 85.3 & 76.1 & 60.7 \\
\oursrow + \ours{} (ours) & \textbf{48.1} & \textbf{79.2} & \textbf{92.7} & \textbf{85.5} & \textbf{86.8} & \textbf{78.9} & \textbf{63.0} \\
\bottomrule
\end{tabular}
\end{adjustbox}
\end{table*}

\subsection{Experimental Setup}
\label{subsec:exp_setup}

We evaluate \ours{} on Qwen2.5-VL-3B, Qwen2.5-VL-7B, and InternVL3.5-8B~\citep{qwen25vl2025,internvl35_2025}. Within each backbone, all RL methods share the initial checkpoint, 35\% ViRL39K~\citep{wang2025vlrethinker} / 65\% verifier-filtered Video-R1~\citep{videor1_2025} curriculum, verifier, rollout group size, and 200-update budget. Baselines are GRPO~\citep{shao2024deepseekmath}, DAPO~\citep{dapo2025}, VPPO~\citep{vppo2025}, and Video-R1~\citep{videor1_2025}; on Qwen2.5-VL-7B we additionally train PEPO~\citep{pepo2026} under the same setting (\Cref{app:pepo_atrl}). Image benchmarks are MME~\citep{mme2023}, MMVP~\citep{mmvp2024}, MMBench~\citep{mmbench2023}, MMStar~\citep{mmstar2024}, CV-Bench~\citep{tong2024cambrian}, POPE~\citep{li2023pope}, RefCOCO~\citep{yu2016modeling}, Geometry3K~\citep{lu2021intergps}, TextVQA~\citep{singh2019towards}, DocVQA~\citep{mathew2021docvqa}, AI2D~\citep{kembhavi2016diagram}, ChartQA~\citep{chartqa2022}, MathVista~\citep{mathvista2023}, and LogicVista~\citep{logicvista2024}. Video benchmarks are Video-MME~\citep{videomme2024}, TempCompass~\citep{tempcompass2024}, NExT-QA~\citep{nextqa2021}, Video-MMMU~\citep{videommmu2025}, LongVideoBench~\citep{longvideobench2024}, and VSI-Bench~\citep{vsibench2024}. Protocols and implementation details are in \Cref{app:evaluation_protocol,app:exp_details}.

\subsection{Main Results}
\label{subsec:main_results}

Across the image and video suites in \Cref{tab:image_general}, \Cref{tab:image_reasoning} and \Cref{tab:video_main}, \ours{} improves most consistently on benchmarks that require visually grounded reasoning rather than broad recognition alone. Video-side evaluation suggests that reward landing is especially useful when relevant evidence is temporally sparse and only a few response positions carry the main decision burden. NExT-QA is the exception on the Qwen backbones, where every RL method falls below the base model, a regression also reported for GRPO on Qwen2.5-VL-7B~\citep{ral2026}; even there \ours{} drops the least, and on InternVL3.5-8B it gains the most. Across the full suite, the results support redistributing outcome-level advantage to visually grounded and decision-bearing tokens without task-specific supervision.

\begin{table*}[t]
\centering
\tsz
\setlength{\tabcolsep}{5pt}
\renewcommand{\arraystretch}{1.06}
\caption{\textbf{Video benchmarks.} Results cover short-video QA, temporal reasoning, knowledge-intensive video QA, long-video understanding, and spatial memory. Higher is better; \textbf{bold} marks the best result within each backbone.}
\label{tab:video_main}
\begin{adjustbox}{max width=\textwidth}
\begin{tabular}{@{}lcccccc@{}}
\toprule
\textbf{Model and method} & \textbf{Video-MME} & \textbf{TempCompass} & \textbf{NExT-QA} & \textbf{Video-MMMU} & \textbf{LongVideoBench} & \textbf{VSI-Bench} \\
\midrule
\tgroup{7}{Qwen2.5-VL-3B} \\
Base model & 58.4 & 62.9 & \textbf{67.7} & 31.3 & 52.2 & 22.3 \\
+ GRPO & 58.2 & 63.1 & 63.1 & 32.4 & 53.1 & 23.5 \\
+ DAPO & 59.0 & 63.8 & 65.8 & 33.3 & 53.6 & 24.4 \\
+ Video-R1 & 58.6 & 63.4 & 64.7 & 32.6 & 52.4 & 22.7 \\
\oursrow + \ours{} (ours) & \textbf{59.5} & \textbf{64.8} & 67.2 & \textbf{33.9} & \textbf{53.7} & \textbf{25.0} \\
\tgroup{7}{Qwen2.5-VL-7B} \\
Base model & 60.1 & 69.0 & \textbf{72.5} & 45.9 & 55.7 & 30.8 \\
+ GRPO & 60.9 & 69.4 & 66.0 & 47.2 & 57.1 & 31.9 \\
+ DAPO & 61.4 & 69.9 & 68.6 & 48.0 & 58.3 & 32.5 \\
+ Video-R1 & 61.8 & 68.5 & 66.5 & 47.6 & 55.4 & 31.8 \\
\oursrow + \ours{} (ours) & \textbf{62.2} & \textbf{72.0} & 71.7 & \textbf{48.3} & \textbf{59.0} & \textbf{33.0} \\
\tgroup{7}{InternVL3.5-8B} \\
Base model & 64.5 & 68.2 & 74.8 & 48.1 & 61.3 & 43.1 \\
+ GRPO & 65.7 & 66.1 & 75.2 & 49.4 & 62.7 & 44.3 \\
+ DAPO & \textbf{67.0} & 68.5 & 76.4 & 50.8 & \textbf{63.2} & 45.2 \\
+ Video-R1 & 65.3 & 68.7 & 73.1 & 49.6 & 61.3 & 43.8 \\
\oursrow + \ours{} (ours) & \textbf{67.0} & \textbf{70.1} & \textbf{78.9} & \textbf{51.2} & 62.7 & \textbf{45.9} \\
\bottomrule
\end{tabular}
\end{adjustbox}
\end{table*}

\Needspace{20\baselineskip}
\begin{wrapfigure}[17]{r}{0.31\textwidth}
    \centering
    \includegraphics[width=\linewidth]{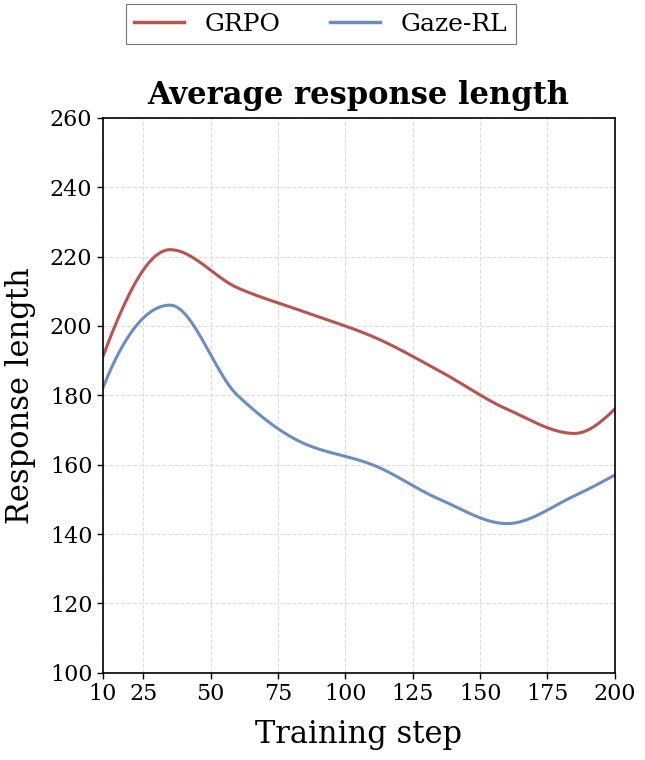}
    \caption{\textbf{Average response length.}}
    \label{fig:response_length}
\end{wrapfigure}

\noindent\textbf{Training dynamics.} \Cref{fig:response_length} shows that the average response length of \ours{} decreases during training and stays below that of GRPO. Details are given in \Cref{app:training_dynamics}.

\FloatBarrier
\subsection{Reward Landing, Visual Evidence, and Ablations}
\label{subsec:response_reward_distribution}

\begin{figure}[t]
    \centering
    \includegraphics[width=1\linewidth]{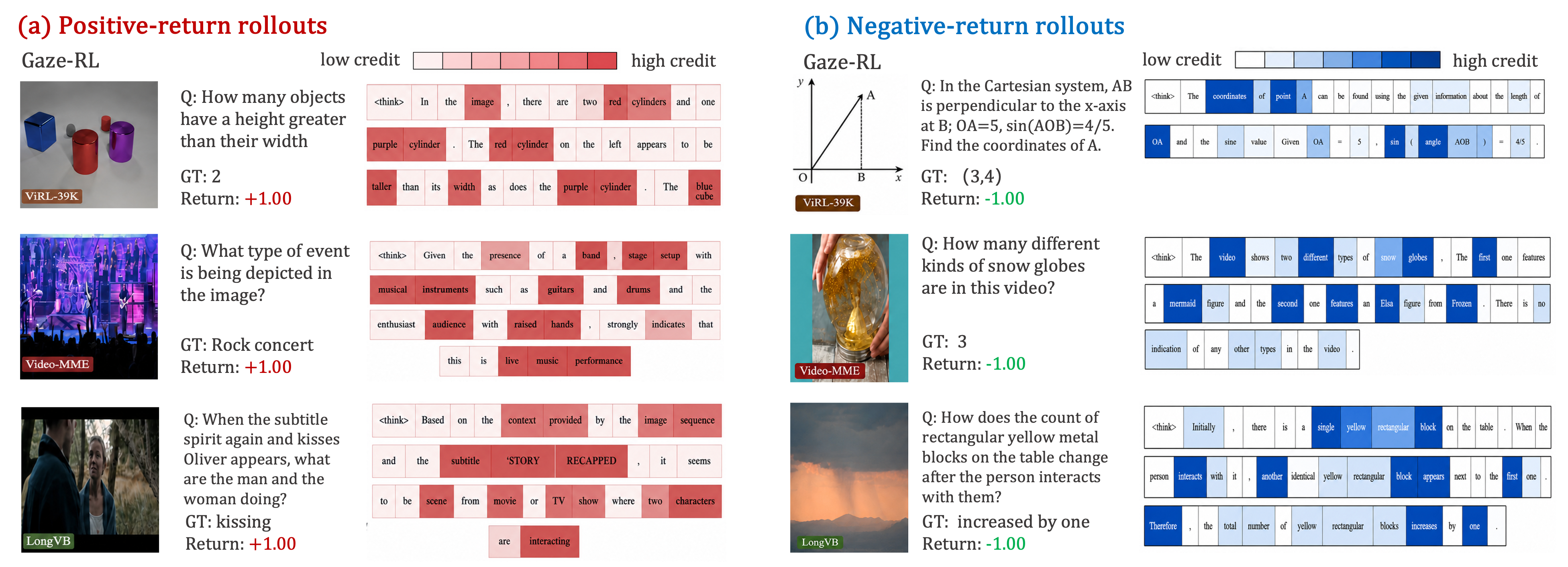}
    \caption{\textbf{Response-side reward landing on diagnostic rollouts.} Red and blue heatmaps show positive- and negative-return examples; darker cells indicate larger landing weight \(\varphi_t\).}
    \label{fig:response_landing_patterns}
\end{figure}

We visualize response-side credit in \Cref{fig:response_landing_patterns}: \ours{} places larger landing weight on visually grounded entities, numerical comparisons, relation or temporal cues, and final decision spans than on filler or formatting tokens. Negative-return examples remain interpretable: \(\varphi_t\) identifies where the update lands, while the sign of \(\widehat A\) determines whether those positions are reinforced or discouraged.

\begin{figure}[t]
    \centering
    \includegraphics[width=1\linewidth]{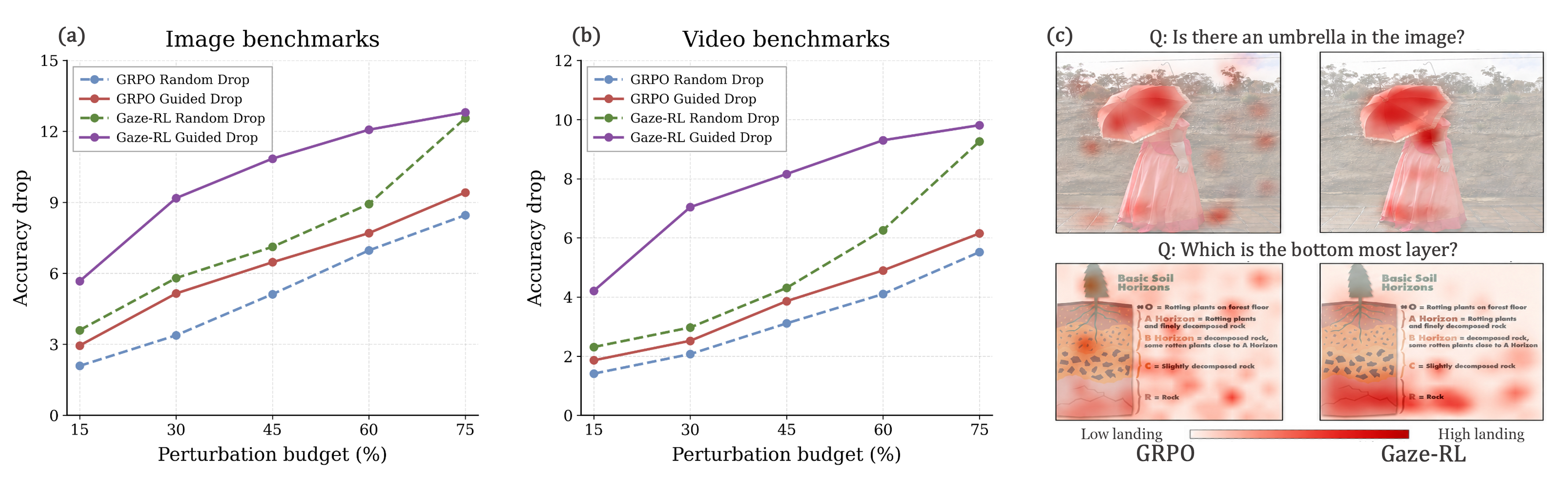}
    \caption{\textbf{Visual evidence sensitivity.} Guided drop masks visual cells ranked by the projected landing field; random drop masks the same budget uniformly. Larger accuracy drops indicate more necessary evidence. (c) Projected landing field of the GRPO and \ours{} checkpoints on two examples.}
    \label{fig:perturbation_and_attention}
\end{figure}

We test whether response-side landing points to visual evidence that affects prediction. \Cref{fig:perturbation_and_attention} visualizes guided and random dropping on the trained checkpoints under matched masking budgets, alongside examples of question-relevant regions. Cross-checkpoint drop magnitudes also depend on each model's masking robustness, so they alone do not isolate ranking quality.

\begin{figure}[!htbp]
    \centering
    \includegraphics[width=1\linewidth]{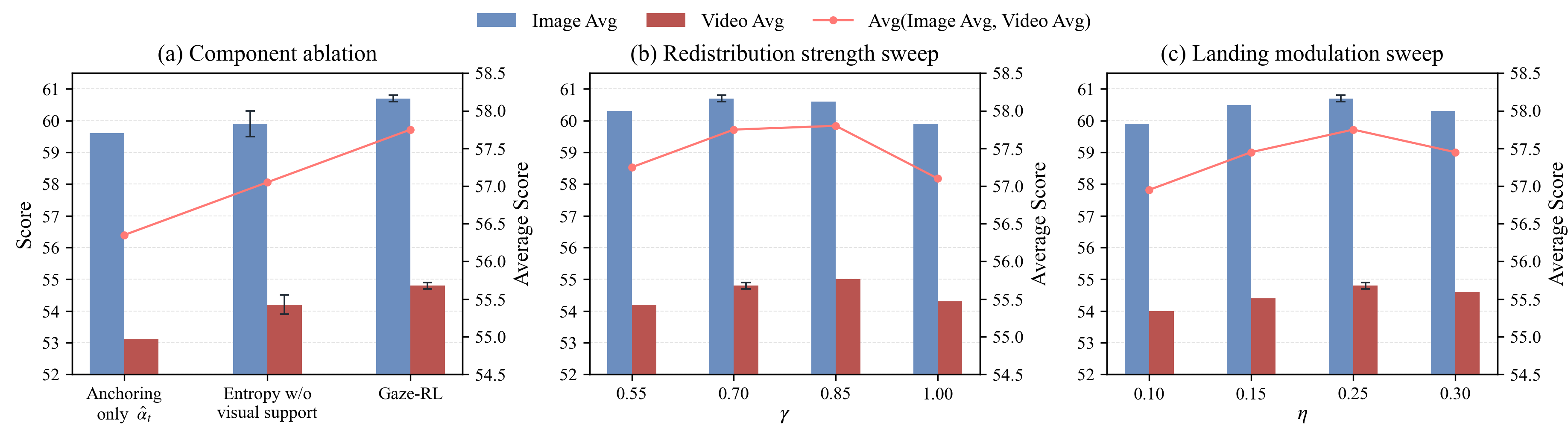}
    \caption{\textbf{Ablation and sweep on Qwen2.5-VL-7B.} Bars report Image Avg and Video Avg; the line reports their mean. Bars with error bars give the mean and standard deviation over three matched seeds: entropy without visual support and the default \ours{} configuration. The other bars are single runs. Image Avg averages MMVP, Geometry3K, MathVista, and LogicVista; Video Avg averages Video-MME, TempCompass, Video-MMMU, LongVideoBench, and VSI-Bench.}
    \label{fig:ablation_and_sweep}
\end{figure}

\Cref{fig:ablation_and_sweep} shows that the full landing rule is stronger than using visual anchoring alone or entropy without visual support. The sweeps favor intermediate settings: performance peaks near the default, with \(\gamma=0.85\) on par with \(\gamma=0.70\), and declines toward weak redistribution, which approaches uniform GRPO-style credit. Across the \(\eta\) sweep the largest possible ratio between two weights, \(e^{2\eta}\), ranges only from \(1.22\) to \(1.82\), so these settings differ only in the strength of a bounded modulation. The benchmark averages are defined in \Cref{app:ablation_average}.

\FloatBarrier
\subsection{Zero-shot Transfer Probe on Image Editing}
\label{subsec:editing_probe}

\begin{figure}[H]
    \centering
    \includegraphics[width=\textwidth]{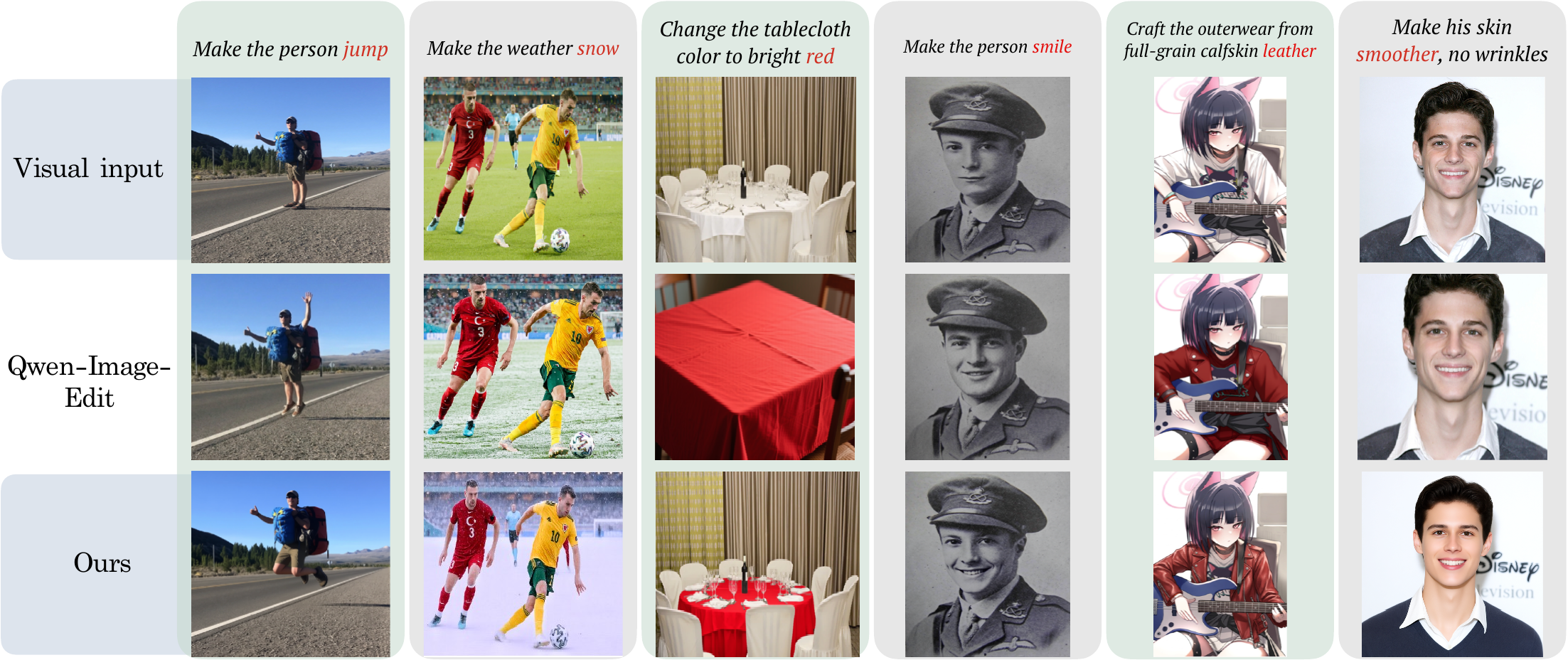}
    \caption{\textbf{Image-editing transfer probe.} We replace the VLM inside Qwen-Image-Edit with the \ours{}-trained Qwen2.5-VL-7B-Instruct while keeping the editing backbone fixed.}
    \label{fig:edit_generation}
\end{figure}

\Needspace{9\baselineskip}
\begin{wraptable}[8]{r}{0.43\textwidth}
\centering
\tsz
\setlength{\tabcolsep}{5pt}
\renewcommand{\arraystretch}{1.06}
\vspace{-12pt}
\caption{\textbf{GEdit-Bench-EN with a fixed editor.} Only the VLM inside Qwen-Image-Edit is replaced.}
\label{tab:gedit}
\begin{tabular}{@{}lc@{}}
\toprule
\textbf{VLM in the editor} & \textbf{Overall} \\
\midrule
Qwen2.5-VL-7B (original) & 7.54 \\
+ GRPO & 7.55 \\
\oursrow + \ours{} (ours) & \textbf{7.60} \\
\bottomrule
\end{tabular}
\end{wraptable}

\Cref{fig:edit_generation} and \Cref{tab:gedit} probe transfer beyond benchmark QA. We replace only the instruction-understanding VLM inside Qwen-Image-Edit~\citep{qwenimage2025} and keep the editing backbone fixed. Qualitatively, the examples suggest improved grounding for local appearance, material, facial attribute, color, weather, and action edits. Quantitatively, on the English instructions of GEdit-Bench~\citep{step1xedit2025}, the \ours{}-trained VLM gives the highest overall score, whereas GRPO training leaves the score nearly unchanged. Since no editing-specific reward is used, we treat this probe as secondary evidence of transfer rather than a primary result. Additional setup details are given in \Cref{app:editing_probe}.

\FloatBarrier
\section{Conclusion and Discussion}
\label{sec:conclusion}

\paragraph{Conclusion.}
This paper revisits multimodal reinforcement learning through a simple but overlooked question: when a verifier rewards an entire response, where should that reward be absorbed inside the generated trajectory? We show that the usual uniform broadcast of completion-level advantage is mismatched with multimodal reasoning, where visual evidence is used sparsely and only a few response positions often mediate the final decision. To address this mismatch, we propose \ours{}, a representation-driven policy optimization method that builds a mean-one reward landing field from internal cross-modal geometry and local predictive uncertainty. The method preserves the total credit of each rollout while reallocating update strength toward positions that are visually anchored and decision-bearing. Across image and video benchmarks, \ours{} improves both general understanding and reasoning-heavy performance. These results suggest that multimodal RL should not only ask whether a trajectory is rewarded, but also learn where that reward should land.

\paragraph{Discussion.}
The main implication is that credit localization should be treated as a core design axis in multimodal post-training. Current methods often improve data, rewards, or optimization stability, but still allow the learning signal to diffuse across long textual continuations that do not equally use visual evidence. \ours{} points to a complementary direction: use the model's own internal representations to identify where visual evidence and consequential uncertainty meet, without adding process labels, external perception modules, or architectural changes. This suggests that the structure of the policy update itself can be a source of grounded reasoning improvement, not merely the reward model or the training data. This perspective is especially relevant for long-video and complex visual reasoning, where decisive evidence may be temporally sparse, locally expressed, and easily diluted by fluent but weakly grounded reasoning text. Future work could extend reward landing beyond token-level fields toward object-, region-, or event-level credit allocation, and combine internal landing signals with stronger verifiers or memory mechanisms for longer-horizon multimodal reasoning.

\paragraph{Limitations.}
The method depends on the quality of the backbone's visual-text representations and of the verifier that supplies the rollout-level advantage. It cannot correct an erroneous outcome reward or missing visual evidence; these limitations and further directions are discussed in Appendices~\ref{app:limitations} and~\ref{app:future_work}.

\renewcommand{\bibfont}{\small}
\setlength{\bibsep}{2pt plus 1pt}
\bibliography{references}
\bibliographystyle{plainnat}

\clearpage
\appendix
\section{Method Details and Derivations}
\label{app:method_details}

This appendix gives the full construction of the latent reward landing field used by \ours{}.
We follow the notation of \Cref{sec:prelim,sec:method}.
For a rollout \(\Yc=(y_1,\ldots,y_T)\), the verifier produces a completion-level advantage \(\widehat A\), while \ours{} constructs a rollout-dependent field over response positions and converts \(\widehat A\) into landed advantages \(\widetilde A_t\).
The construction is post-hoc with respect to the sampled rollout: it uses the policy's internal hidden states and rollout-time predictive distributions, but introduces no additional reward model, process supervision, auxiliary perception module, or architectural modification.

\subsection{Notation summary}
\label{app:notation_summary}

\Cref{tab:notation_summary} summarizes the main quantities used in the method and appendix.

\begin{table}[h]
\centering
\tsz
\setlength{\tabcolsep}{5pt}
\renewcommand{\arraystretch}{1.06}
\caption{\textbf{Notation.} Quantities used in the reward landing construction.}
\label{tab:notation_summary}
\begin{adjustbox}{max width=\linewidth}
\begin{tabular}{@{}ll@{}}
\toprule
\textbf{Symbol} & \textbf{Meaning} \\
\midrule
\(\Vc=(v_1,\ldots,v_M)\) & visual embeddings or packed visual memory elements \\
\(\Pc\) & textual prompt \\
\(\Yc=(y_1,\ldots,y_T)\) & generated response \\
\(s_t=(\Vc,\Pc,\Yc_{<t})\) & autoregressive state before generating \(y_t\) \\
\(\widehat A\) & group-normalized completion-level advantage \\
\(\mathcal L\) & selected transformer layers used for internal geometry \\
\(h^{\mathrm{txt}}_{\ell,t}\) & pre-action response state predicting \(y_t\) at layer \(\ell\) \\
\(h^{\mathrm{vis}}_{\ell,m}\) & hidden state of visual memory element \(v_m\) at layer \(\ell\) \\
\(\alpha_t,\widehat\alpha_t\) & raw and rollout-normalized visual anchoring scores \\
\(H_t\) & rollout-time predictive entropy at position \(t\) \\
\(z_t\) & layer-averaged, state-normalized response state \\
\(K_{\tau t}\) & causal support kernel over predecessors \(\tau<t\) \\
\(\zeta_t\) & recursive visual support of position \(t\) \\
\(b_t\) & supported-branching score \\
\(e_t,\tilde e_t\) & log-evidence and its centered version \\
\(\varphi_t\) & constructed mean-one landing field \\
\(\Phi_\gamma(t)\) & blended landing field with landing strength \(\gamma\) \\
\(\widetilde A_t\) & landed token-level advantage \\
\bottomrule
\end{tabular}
\end{adjustbox}
\end{table}

\subsection{Credit preservation of mean-one landing fields}
\label{app:credit_preservation}

The defining constraint of a reward landing field is
\[
    \Phi(t)\ge 0,
    \qquad
    \frac{1}{T}\sum_{t=1}^{T}\Phi(t)=1,
    \qquad
    \widetilde A_t=\Phi(t)\widehat A.
\]
This constraint is important because \ours{} is designed to redistribute the existing completion-level advantage rather than amplify or attenuate it.
Indeed,
\begin{equation}
    \sum_{t=1}^{T}\widetilde A_t
    =
    \sum_{t=1}^{T}\Phi(t)\widehat A
    =
    T\widehat A.
    \label{eq:app_total_credit_preservation}
\end{equation}
Thus the average landed advantage over response positions is still \(\widehat A\).
When \(\Phi(t)\equiv 1\), the landed advantage reduces to the usual uniform assignment \(\widetilde A_t=\widehat A\), recovering the GRPO-style update.

The constructed field \(\varphi_t\) in \Cref{eq:landing_field_softmax} is also mean-one by construction.
Let
\[
    q_t=\exp\!\bigl(\eta\tanh\tilde e_t\bigr),
    \qquad
    Z=\sum_{\tau=1}^{T}q_\tau.
\]
Then \(\varphi_t=Tq_t/Z\), so
\begin{equation}
    \frac{1}{T}\sum_{t=1}^{T}\varphi_t
    =
    \frac{1}{T}\sum_{t=1}^{T}T\frac{q_t}{Z}
    =
    1.
    \label{eq:app_phi_mean_one}
\end{equation}
The blended field is
\(\Phi_\gamma(t)=(1-\gamma)+\gamma\varphi_t\), with \(\gamma\in[0,1]\).
Because both the uniform field and \(\varphi_t\) are mean-one,
\begin{equation}
    \frac{1}{T}\sum_{t=1}^{T}\Phi_\gamma(t)
    =
    (1-\gamma)
    +
    \gamma\frac{1}{T}\sum_{t=1}^{T}\varphi_t
    =
    1.
    \label{eq:app_blended_mean_one}
\end{equation}
Therefore, for every \(\gamma\), the total rollout credit remains \(T\widehat A\), while \(\gamma\) controls how far the update moves away from uniform token-level credit.

\subsection{Rollout-wise scalar normalization}
\label{app:rollout_norm}

For any rollout-side scalar sequence \(q_{1:T}\), the operator \(\operatorname{Norm}_{1:T}\) used in \Cref{eq:visual_anchor_score} applies stabilized min-max scaling over response positions:
\begin{equation}
    \operatorname{Norm}_{1:T}(q_t)
    =
    \frac{
        q_t-\min_{\tau\in[T]}q_\tau
    }{
        \max_{\tau\in[T]}q_\tau-\min_{\tau\in[T]}q_\tau+\varepsilon_{\mathrm{norm}}
    },
    \qquad
    \varepsilon_{\mathrm{norm}}>0.
    \label{eq:app_rollout_norm}
\end{equation}
The normalization is local to each generated response, rather than shared across prompts or minibatches.
This is useful because raw anchoring scales can vary across inputs, modalities, response lengths, and model backbones.
Rollout-wise scaling keeps only the within-response contrast needed for positionwise credit allocation.
When the score range is nearly flat, the denominator is dominated by \(\varepsilon_{\mathrm{norm}}\) and the normalized contrast of that score remains small.
Up to \(\varepsilon_{\mathrm{norm}}\), the operator is also invariant to positive rescaling, \(\operatorname{Norm}_{1:T}(\lambda q_t)=\operatorname{Norm}_{1:T}(q_t)\) for \(\lambda>0\), so a factor shared by all positions does not change the normalized signal.

\subsection{Stabilized state normalization}
\label{app:state_norm}

The operator \(\operatorname{StateNorm}\) that forms the per-position response state \(z_t\) in \Cref{eq:branch_kernel} applies an \(\ell_2\) normalization with a small additive denominator:
\begin{equation}
    \operatorname{StateNorm}(z)
    =
    \frac{z}{\|z\|_2+\varepsilon_{\mathrm{sup}}},
    \qquad
    \varepsilon_{\mathrm{sup}}>0.
    \label{eq:app_state_norm}
\end{equation}
This prevents division by near-zero norms when a layer-averaged hidden state has very small magnitude.
The role of this normalization is only to make response-state similarity comparable across positions; it does not change the model's forward pass or add a trainable component.

\subsection{Supported-branching coefficient}
\label{app:branch_coeff}

Visual anchoring identifies positions whose hidden states are close to the visual stream, but not every visually anchored token is a consequential decision point.
For example, a token can be visually grounded while its next-token distribution is nearly deterministic.
Supported branching therefore adds a second signal: it gives higher weight to positions where the model faces local predictive uncertainty and where the preceding states supporting that uncertainty are themselves visually anchored.

The local predictive entropy at position \(t\) is
\begin{equation}
    H_t
    =
    \operatorname{Ent}\!\bigl(\pi_{\theta_{\mathrm{old}}}(\cdot\mid s_t)\bigr)
    =
    -\sum_{c\in\Sigma_{\mathrm{tok}}}
    \pi_{\theta_{\mathrm{old}}}(c\mid s_t)
    \log\pi_{\theta_{\mathrm{old}}}(c\mid s_t),
    \label{eq:local_entropy}
\end{equation}
where \(\Sigma_{\mathrm{tok}}\) is the token vocabulary.
The entropy is computed from the rollout-time old policy, matching the policy distribution that generated the response.
High \(H_t\) indicates that the model has multiple plausible continuations at position \(t\), but entropy alone is not sufficient: uncertainty can arise from harmless wording variation or formatting, rather than from a visually meaningful reasoning branch.

The support coefficient \(\zeta_t\) of \Cref{eq:branch_coeff} keeps only the part of this uncertainty that is supported by visual evidence.
It first summarizes each position by \(z_t\) in \Cref{eq:branch_kernel}, the layer average of the pre-action states over the same layer set \(\mathcal L\) used for visual anchoring, normalized by \(\operatorname{StateNorm}\).
The causal kernel \(K_{\tau t}\) then weights each predecessor \(\tau<t\) by response-state similarity with temperature \(\kappa\) and by a log-distance decay \(\lambda_{\mathrm{lag}}\log(1+t-\tau)\).
The similarity term selects earlier response states that are representation-wise close to the current state, while the log-distance penalty prevents very early tokens from dominating purely because of broad semantic similarity.
The kernel is causal because it only uses predecessor positions \(\tau<t\).
In the implementation, all response states are stacked and the kernel is evaluated with batched matrix multiplication and a strict lower-triangular mask. Cached decoder states are shifted so that the state and old-policy entropy at position \(t\) both refer to the pre-action state \(s_t\) that predicts \(y_t\). The complete landing field is computed in FP32 and detached before policy optimization.

In \Cref{eq:branch_coeff}, the first term credits a position for its own anchoring and the second for the support of the predecessors selected by the kernel; since those predecessors are defined the same way, support propagates along chains of related response states.
This is what distinguishes visually supported uncertainty from ordinary language uncertainty: a position whose own state and selected predecessors are weakly anchored relative to the rest of the rollout receives little support, and its entropy does not receive a visual-branching boost.
Support is relative to the rollout through the min-max normalization of \(\alpha_t\), and \(\chi\) sets how much of a position's support comes from the states it draws on.
The first generated token has no predecessor, so \(\zeta_1=(1-\chi)\,\widehat\alpha_1\).
In the implementation, \(\zeta\) is computed for all positions at once by a batched triangular solve of \((I-\chi K^{\top})\zeta=(1-\chi)\widehat\alpha\), and the scores \(b_t\) are floored at \(10^{-12}\) before the logarithm \(e_t=\log b_t\) of \Cref{eq:landing_field_softmax}.
Together, these choices make \(b_t\) large only when a position is both locally uncertain and visually anchored or supported by visually anchored earlier states.

\subsection{Limiting behavior of the support coefficient}
\label{app:limiting_behavior}

This subsection proves the statements summarized at the end of \Cref{subsec:method_objective}.

\paragraph{The support coefficient acts only through its variation.}
A factor \(\lambda>0\) shared by all \(b_t\) adds \(\log\lambda\) to every \(e_t\) and leaves the centered \(\tilde e_t\), and hence \Cref{eq:landing_field_softmax}, unchanged.
Hence any constant coefficient \(\zeta_t\equiv\zeta>0\) yields \(\tilde e_t=\log H_t-\frac1T\sum_\tau\log H_\tau\): the field reduces to one of entropy alone.
The coefficient is therefore a relative gate that re-ranks uncertain positions by their visual support, not a switch that turns entropy off.
The variant \emph{entropy without visual support} in \Cref{fig:ablation_and_sweep} trains exactly this limit.

\paragraph{Bounds of the field.}
For any two positions \(s\) and \(t\),
\begin{equation}
    \frac{\varphi_t}{\varphi_s}=\exp\!\bigl(\eta\,(\tanh\tilde e_t-\tanh\tilde e_s)\bigr)\le e^{2\eta},
    \label{eq:app_weak_bound}
\end{equation}
so the field depends on the relative pattern of \(b_t\) and not on its scale.
Averaging \Cref{eq:app_weak_bound} over \(s\) and using \(\frac1T\sum_s\varphi_s=1\) gives \(e^{-2\eta}\le\varphi_t\le e^{2\eta}\) at every position, and hence
\begin{equation}
    1-\gamma\bigl(1-e^{-2\eta}\bigr)\le\Phi_\gamma(t)\le1+\gamma\bigl(e^{2\eta}-1\bigr).
    \label{eq:app_blend_bounds}
\end{equation}
At the defaults \(\gamma=0.70\) and \(\eta=0.25\), the ratio bound is \(1.65\) and \Cref{eq:app_blend_bounds} gives \(0.72\le\Phi_\gamma(t)\le1.45\).
The ratio bound also limits the landing mass of any set of positions: a set that holds a fraction \(p\) of the positions of a rollout receives at most a fraction \(pr/(pr+1-p)\) of \(\sum_t\varphi_t\), with \(r=e^{2\eta}\); for the highest-weighted \(20\%\) of positions this is \(29.2\%\) at the default \(\eta\).

\subsection{Landed policy objective}
\label{app:landed_objective_detail}

After constructing \(\varphi_t\), \ours{} uses the blended field
\[
    \Phi_\gamma(t)=(1-\gamma)+\gamma\varphi_t,
    \qquad
    \gamma\in[0,1],
\]
and defines the landed advantage
\[
    \widetilde A_t=\Phi_\gamma(t)\widehat A.
\]
The policy objective is the same clipped surrogate used by GRPO, except that each response position receives \(\widetilde A_t\) instead of the uniform \(\widehat A\):
\[
    \mathcal J_{\ours}(\theta)
    =
    \mathbb E_{\mathrm{rollout}}
    \left[
        \frac{1}{T}
        \sum_{t=1}^{T}
        \left(
            \psi_\epsilon(\rho_t(\theta),\widetilde A_t)
            -
            \beta\,\mathrm{KL}_t(\theta)
        \right)
    \right].
\]
All other components are unchanged: the verifier, group-relative advantage computation, clipping radius, and reference-policy penalty are the same as in the corresponding GRPO training run.
Thus the method changes where the update lands, not which trajectories are rewarded or how much total advantage they receive.

The sign of the update is still controlled by \(\widehat A\).
If \(\widehat A>0\), high-\(\Phi_\gamma(t)\) positions receive stronger positive reinforcement.
If \(\widehat A<0\), the same positions receive stronger discouragement.
This is why negative-return examples remain interpretable in the response-side visualizations: the landing field identifies the positions responsible for absorbing the update, while the completion-level advantage determines the update direction.

\section{Experimental Details}
\label{app:exp_details}

\subsection{Training Data and Baselines}
\label{app:training_data_baselines}

The training set is a mixed image-video RL curriculum constructed from ViRL39K~\citep{wang2025vlrethinker} and a verifier-filtered subset of Video-R1~\citep{videor1_2025}. Image data counts for 35\% and video data counts for 65\%. The image portion covers visual logic, chart reasoning, mathematical diagrams, OCR-heavy question answering, and grounding-oriented question answering. The video portion covers short-video understanding, temporal ordering, event localization, and long-video evidence aggregation. For the three backbones in the main tables, GRPO~\citep{shao2024deepseekmath}, DAPO~\citep{dapo2025}, VPPO~\citep{vppo2025}, Video-R1~\citep{videor1_2025}, and \ours{} use the same initial checkpoint, curriculum, verifier, 200-update budget, rollout group size, and hardware within each backbone. Their optimization rules differ, but the training data are not a method-specific advantage.

\subsection{Training Configuration}
\label{app:training_config}

\Cref{tab:training_config} summarizes the main training and rollout settings used in our implementation. We train for a fixed number of RL updates, save checkpoints periodically, and use grouped rollouts for advantage normalization. The generation budget is capped at \(512\) new tokens, which matches the response-length analysis in \Cref{fig:response_length} under the same generation limit. For video inputs, we sample at \(0.5\) fps and cap the number of sampled frames. Image resolution is controlled by the pixel budget of the corresponding backbone.

\begin{table}[t]
\centering
\tsz
\setlength{\tabcolsep}{5pt}
\renewcommand{\arraystretch}{1.06}
\caption{\textbf{Main RL training and rollout configuration.} These settings are shared by the reported \ours{} runs unless otherwise specified.}
\label{tab:training_config}
\begin{adjustbox}{max width=\linewidth}
\begin{tabular}{@{}ll@{}}
\toprule
\textbf{Setting} & \textbf{Value} \\
\midrule
\tgroup{2}{Optimization} \\
Training steps & \(200\) \\
Checkpoint interval & every \(5\) updates \\
Hardware & \(4\) NVIDIA H200 GPUs \\
Gradient accumulation & \(32\) steps \\
Learning rate & \(1.0\times 10^{-6}\) \\
Weight decay & \(0.0\) \\
Warmup ratio & \(0.03\) \\
Learning-rate schedule & cosine \\
Maximum gradient norm & \(1.0\) \\
\tgroup{2}{Rollout} \\
Rollout group size & \(G=8\) \\
Rollout chunk size & \(2\) \\
Maximum new tokens & \(512\) \\
Minimum generated tokens & \(1\) \\
Maximum sequence length & \(5120\) \\
Maximum generation time & \(600\) seconds \\
\bottomrule
\end{tabular}
\end{adjustbox}
\end{table}

\subsection{Optimization and Reward Settings}
\label{app:optimization_reward}

Group advantages follow \Cref{eq:group_adv}, with the population standard deviation over the \(G\) rollouts of a prompt and the additive stabilizer \(\delta=10^{-3}\) in the denominator. A group in which every rollout receives the same reward therefore contributes zero advantage. The reward also contains a graded format term, so a group whose rewards differ only slightly is divided by a correspondingly small standard deviation and can reach large normalized advantages; clipping the normalized advantages to \([-2.5,2.5]\) before landing bounds these groups as well as rare outliers. The PPO-style update uses clipping with \(\epsilon=0.2\) and a reference-policy KL coefficient of \(\beta=0.001\), with the per-token KL term estimated from the sampled token by the non-negative \(k_3\) estimator \(\exp(\log\pi_{\mathrm{ref}}-\log\pi_\theta)-(\log\pi_{\mathrm{ref}}-\log\pi_\theta)-1\). The reward combines answer correctness and response-format quality, with the answer component carrying the dominant weight.

The reward settings are intended to keep the verifier signal outcome-level. The format component only encourages parseable responses, while the answer component determines the main correctness signal. \ours{} does not add token-level rewards; it only redistributes the resulting completion-level advantage across generated response positions.

\section{Compute Resources}
\label{app:compute_resources}

We report the computational resources used for the representative training configuration in \Cref{tab:compute_wallclock}. The numbers are measured for \(200\) RL updates under the same rollout and optimization configuration used in the main experiments. These measurements are intended to characterize the additional overhead of the landing-field computation relative to vanilla GRPO; they should not be interpreted as hardware-independent complexity estimates.

\begin{table}[t]
\centering
\tsz
\setlength{\tabcolsep}{5pt}
\renewcommand{\arraystretch}{1.06}
\caption{\textbf{Compute for \(200\) RL training updates.} \ours{} has nearly the same runtime as vanilla GRPO because rollout generation and policy optimization dominate the total training time. The small memory increase comes from storing selected hidden states, rollout-time logits, and intermediate landing-field statistics.}
\label{tab:compute_wallclock}
\begin{adjustbox}{max width=\linewidth}
\begin{tabular}{@{}lcc@{}}
\toprule
\textbf{Quantity} & \textbf{Vanilla GRPO} & \textbf{\ours{}} \\
\midrule
Wall-clock time & \(2.1\) days & \(2.1\) days \\
Peak allocated VRAM & \(77.7\) GB & \(78.1\) GB \\
\bottomrule
\end{tabular}
\end{adjustbox}
\end{table}

\subsection{Complexity and Memory Overhead}
\label{app:complexity}

The additional computation of \ours{} comes from constructing the landing field during training. The method does not modify the model architecture and does not require an auxiliary perception network. It uses hidden states from selected transformer layers, rollout-time logits for entropy, and the visual memory already produced by the multimodal policy.

For a rollout of length \(T\), hidden dimension \(d\), selected layer count \(|\mathcal L|\), and visual memory size \(M\), visual anchoring with pooled visual prototypes costs \(O(|\mathcal L|Td)\) after the visual prototype of each selected layer is formed. The supported-branching kernel in its full causal form costs \(O(T^2d)\), since each response position compares to its predecessors. This cost is incurred only during training-time landing-field computation. In long-response settings, the same causal support computation can be restricted to a finite predecessor window or approximated with a top-\(k\) predecessor set without changing the definition of the landing field.

The memory overhead is mainly due to storing selected-layer response hidden states, selected-layer visual hidden states, and rollout-time logits. In our implementation, we control memory through chunked rollout generation and forward computation. The rollout chunk size is \(2\), and the forward chunk size is \(1\). Since the landing field is only used for training and analysis, inference with the trained policy has the same architecture and does not require landing-field computation.

\subsection{Evaluation Protocol}
\label{app:evaluation_protocol}

For image-side evaluation, we report results on MME~\citep{mme2023}, MMVP~\citep{mmvp2024}, MMBench~\citep{mmbench2023}, MMStar~\citep{mmstar2024}, CV-Bench~\citep{tong2024cambrian}, POPE~\citep{li2023pope}, RefCOCO~\citep{yu2016modeling}, Geometry3K~\citep{lu2021intergps}, TextVQA~\citep{singh2019towards}, DocVQA~\citep{mathew2021docvqa}, AI2D~\citep{kembhavi2016diagram}, ChartQA~\citep{chartqa2022}, MathVista~\citep{mathvista2023}, and LogicVista~\citep{logicvista2024}. For video-side evaluation, we report results on Video-MME~\citep{videomme2024}, TempCompass~\citep{tempcompass2024}, NExT-QA~\citep{nextqa2021}, Video-MMMU~\citep{videommmu2025}, LongVideoBench~\citep{longvideobench2024}, and VSI-Bench~\citep{vsibench2024}. We follow the official evaluation protocol of each benchmark whenever available, and within each backbone all methods are evaluated with the same prompts, decoding, and, for videos, the same frame sampling. Accuracy is used unless the benchmark defines a different official score; MME follows the original summed score, and RefCOCO reports the average over the standard referring-expression splits used in our evaluation.

\subsection{Training-Dynamics Measurement}
\label{app:training_dynamics}

For \Cref{fig:response_length}, average response length is measured on the same diagnostic rollout set across training checkpoints and counts generated response tokens only. The curves are used only to analyze the evolution of response length; they are not used as training objectives.

\subsection{Visual Evidence Perturbation}
\label{app:visual_perturbation}

For \Cref{fig:perturbation_and_attention}, we project response-side landing mass onto visual memory elements using the model's internal cross-modal similarity. For images, visual memory elements are pooled into spatial cells; for videos, they are pooled into temporal memory units after frame packing. Guided drop masks the highest-ranked cells under the projected landing score, while random drop masks the same number of cells uniformly. We sweep the perturbation budget and re-run the model with the perturbed input. The reported curve measures the absolute accuracy drop relative to the full visual input.

This perturbation is used only for analysis. It is not used during training and does not provide supervision to the landing field. The comparison between guided drop and random drop tests whether the projected landing field identifies visual evidence that is necessary for the model's prediction. A larger guided-drop accuracy decrease indicates that the selected visual cells are more causally important for the answer.

\subsection{Ablation Average Definition}
\label{app:ablation_average}

The ablation and hyperparameter sweep in \Cref{fig:ablation_and_sweep} are conducted on Qwen2.5-VL-7B. Image Avg is the arithmetic mean of MMVP, Geometry3K, MathVista, and LogicVista. Video Avg is the arithmetic mean of Video-MME, TempCompass, Video-MMMU, LongVideoBench, and VSI-Bench. These subsets are selected because they emphasize visually grounded reasoning and temporal evidence usage.

\subsection{Image-Editing Transfer Probe}
\label{app:editing_probe}

For \Cref{subsec:editing_probe}, we use the image-editing system only as a transfer probe. The image-editing backbone of Qwen-Image-Edit~\citep{qwenimage2025} is kept fixed, and only its instruction-understanding VLM, Qwen2.5-VL-7B-Instruct, is replaced by the GRPO- or \ours{}-trained checkpoint of the same backbone; no editing-specific reward or fine-tuning is used. The qualitative prompts in \Cref{fig:edit_generation} cover local action changes, weather changes, object color changes, facial attributes, and material-specific edits, and involve no human or automatic scoring. The quantitative results in \Cref{tab:gedit} use the English instructions of GEdit-Bench~\citep{step1xedit2025} and report its overall score under the official evaluation protocol, with the same editing settings for all three VLMs. The purpose of this probe is to test whether reward landing improves instruction grounding in a generation pipeline outside the main QA benchmarks.

\section{Additional Controlled Analyses}
\label{app:controlled_analyses}

\subsection{Comparison with PEPO}
\label{app:pepo_atrl}

PEPO~\citep{pepo2026} reweights token advantages by combining response--image hidden-state similarity with token entropy.
We trained it on Qwen2.5-VL-7B-Instruct with the same initialization and evaluated it with the protocol of \Cref{tab:image_general,tab:image_reasoning}.
The base model, GRPO, and \ours{} rows in \Cref{tab:pepo_atrl} are repeated from those tables.

\begin{table}[H]
\centering
\tsz
\setlength{\tabcolsep}{5pt}
\renewcommand{\arraystretch}{1.06}
\caption{\textbf{Comparison with PEPO under the main training setting.} All methods start from Qwen2.5-VL-7B-Instruct and share the training mix, verifier, rollout group size, and update budget. \textbf{Bold} marks the best result in each column.}
\label{tab:pepo_atrl}
\begin{adjustbox}{max width=\linewidth}
\begin{tabular}{@{}lccc@{}}
\toprule
\textbf{Model and method} & \textbf{MMVP} & \textbf{MMStar} & \textbf{AI2D} \\
\midrule
\tgroup{4}{Qwen2.5-VL-7B} \\
Base model & 76.3 & 62.6 & 83.2 \\
+ GRPO & 76.7 & 63.2 & 84.2 \\
+ PEPO & 76.7 & 63.6 & 82.4 \\
\oursrow + \ours{} (ours) & \textbf{78.7} & \textbf{64.4} & \textbf{85.0} \\
\bottomrule
\end{tabular}
\end{adjustbox}
\end{table}

\ours{} is best on all three benchmarks under this matched setting, while PEPO falls below GRPO on AI2D.

\subsection{Negative-Advantage Landing}
\label{app:negative_advantage}

Mean-one normalization preserves scalar rollout-credit mass but does not prove that every negative update penalizes the correct cause of an error. By \Cref{eq:app_blend_bounds}, the defaults \(\gamma=0.70\) and \(\eta=0.25\) keep \(0.72\le\Phi_\gamma(t)\le1.45\), so every token of a negative-advantage rollout receives at least about \(72\%\) of the uniform GRPO penalty before clipping, and the landing field only modulates that penalty by a bounded factor.

\section{Limitations}
\label{app:limitations}

\paragraph{Dependence on internal representation quality.}
\ours{} assumes that the model's internal cross-modal geometry contains usable information about where visual evidence is being used. This assumption is reasonable for strong multimodal backbones, but it can fail when the backbone itself has weak visual-text binding, poor OCR grounding, or unstable video memory. In such cases, the landing field may shift credit toward positions that appear visually anchored in representation space but do not correspond to genuinely relevant evidence. Thus, \ours{} should be viewed as a way to better allocate an existing learning signal inside a capable policy, rather than as a replacement for strong multimodal pretraining or reliable visual representation learning.

\paragraph{Outcome reward remains the source of update direction.}
The landing field determines where the completion-level advantage is absorbed, but it does not determine whether the rollout should be rewarded or penalized. The sign and magnitude of the update still come from the verifier-derived advantage. If the verifier assigns an incorrect or biased reward, \ours{} only redistributes the resulting update across positions and cannot correct the reward itself. This limitation is inherent to methods that redistribute outcome-level rewards without changing the reward model. More reliable verifiers, uncertainty-aware reward models, or calibrated rejection mechanisms remain important for robust multimodal RL.

\section{Future Work}
\label{app:future_work}

\paragraph{From token landing to structured visual credit.}
A natural next step is to extend reward landing from response-token fields to structured visual credit fields. Instead of only assigning update mass to generated positions, future methods could jointly infer which response tokens, image regions, objects, video frames, and temporal events should absorb the learning signal. This would make the connection between reasoning text and visual evidence more explicit, especially for long-video understanding, spatial reasoning, and multi-object interaction.

\paragraph{Joint modeling of reward quality and landing confidence.}
In the current formulation, the verifier supplies the trajectory-level advantage and the landing field supplies positionwise allocation. Future work could model these two uncertainties jointly: a rollout with uncertain reward, weak visual anchoring, or flat landing contrast should induce a more conservative update. Such a direction could combine reward calibration, verifier ensembles, and internal landing confidence, allowing the optimizer to decide not only where reward should land, but also how strongly it should be trusted.

\paragraph{More efficient landing computation.}
The causal support computation can be made more scalable through sparse predecessor selection, windowed kernels, low-rank state summaries, or online updates during generation. These approximations are especially relevant for long-context video reasoning, where both generated responses and visual memories can become large. An efficient implementation would make reward landing easier to combine with long-video memory, retrieval-augmented visual context, and streaming multimodal policies. Combining these mechanisms with stronger verifiers is another direction for longer-horizon multimodal reasoning.

\paragraph{Beyond verifiable QA.}
Reward landing is most directly evaluated here under verifiable image and video QA settings, but the underlying idea is not tied to multiple-choice or short-answer evaluation. Future work could test the same principle in instruction-guided image editing, interactive visual dialogue, multimodal planning, robotics, and tool-augmented visual reasoning. These settings may require richer rewards and stronger safety constraints, but they also make the core question more important: when a long multimodal trajectory succeeds or fails, which parts of the trajectory should actually receive the learning signal?

\section{Existing Assets and Licenses}
\label{app:asset_licenses}

\Cref{tab:asset_licenses} lists the main existing assets used in this work. We use public model releases, public datasets, public benchmarks, and standard open-source software libraries. For each asset, we cite the corresponding paper or project and follow the license or terms specified by the official release. We do not introduce a new dataset in this paper.

\begin{table}[t]
\centering
\caption{\textbf{Existing assets used in this work.} Licenses and terms are taken from the official model, dataset, or project release pages when available.}
\label{tab:asset_licenses}
\tsz
\setlength{\tabcolsep}{5pt}
\renewcommand{\arraystretch}{1.06}
\begin{tabularx}{\linewidth}{@{}>{\hsize=1.4\hsize\raggedright\arraybackslash}X >{\hsize=.7\hsize\raggedright\arraybackslash}X >{\hsize=1.05\hsize\raggedright\arraybackslash}X >{\hsize=.85\hsize\raggedright\arraybackslash}X@{}}
\toprule
\textbf{Asset} & \textbf{Type} & \textbf{Usage in this paper} & \textbf{License / terms} \\
\midrule
Qwen2.5-VL-3B / Qwen2.5-VL-7B~\citep{qwen25vl2025}
& Model backbone 
& Main multimodal policy backbones 
& Apache-2.0 official release \\
InternVL3.5-8B~\citep{internvl35_2025}
& Model backbone 
& Main multimodal policy backbone 
& Apache-2.0 official release \\
ViRL39K~\citep{wang2025vlrethinker} 
& Training dataset 
& Image-side RL curriculum 
& MIT official dataset release \\
Video-R1~\citep{videor1_2025}
& Training dataset / baseline 
& Video-side RL curriculum and video RL baseline 
& Apache-2.0 official dataset release \\
Qwen-Image-Edit~\citep{qwenimage2025}
& Image-editing model 
& Image-editing transfer probe 
& Apache-2.0 official release \\
GEdit-Bench~\citep{step1xedit2025}
& Evaluation benchmark 
& Quantitative image-editing transfer probe 
& Public benchmark terms \\
MME~\citep{mme2023}, MMVP~\citep{mmvp2024}, MMBench~\citep{mmbench2023}, MMStar~\citep{mmstar2024}
& Evaluation benchmarks 
& Image-side general and diagnostic evaluation 
& Public benchmark terms \\
CV-Bench~\citep{tong2024cambrian}, POPE~\citep{li2023pope}, RefCOCO~\citep{yu2016modeling}
& Evaluation benchmarks 
& Visual perception, hallucination, and grounding evaluation 
& Public benchmark terms \\
Geometry3K~\citep{lu2021intergps}, TextVQA~\citep{singh2019towards}, DocVQA~\citep{mathew2021docvqa}, AI2D~\citep{kembhavi2016diagram}
& Evaluation benchmarks 
& OCR-heavy, diagram, and visual reasoning evaluation 
& Public benchmark terms \\
ChartQA~\citep{chartqa2022}, MathVista~\citep{mathvista2023}, LogicVista~\citep{logicvista2024}
& Evaluation benchmarks 
& Chart, mathematical, and logical reasoning evaluation 
& Public benchmark terms \\
Video-MME~\citep{videomme2024}, TempCompass~\citep{tempcompass2024}, NExT-QA~\citep{nextqa2021}
& Evaluation benchmarks 
& Video understanding and temporal reasoning evaluation 
& Public benchmark terms \\
Video-MMMU~\citep{videommmu2025}, LongVideoBench~\citep{longvideobench2024}, VSI-Bench~\citep{vsibench2024}
& Evaluation benchmarks 
& Knowledge-intensive, long-video, and spatial memory evaluation 
& Public benchmark terms \\
PyTorch, Transformers, vLLM 
& Software libraries 
& Training, inference, and rollout generation 
& Open-source software licenses \\
\bottomrule
\end{tabularx}
\end{table}

For benchmarks and software packages whose licenses are maintained by their official repositories, we follow the corresponding official terms of use. The released code accompanying this paper is intended for research use and should be used together with the licenses of the underlying models, datasets, and software dependencies.

\section{Broader Impact}
\label{app:broader_impact}

This work studies how to improve the allocation of existing reinforcement-learning signals in multimodal reasoning models. The potential positive impact is improved grounding: by shifting policy updates toward response positions that are visually anchored and decision-bearing, the method may reduce weakly grounded reasoning and improve evidence sensitivity. This can benefit applications that require faithful multimodal understanding, such as visual question answering, educational diagram reasoning, chart analysis, accessibility tools, and long-video retrieval or summarization.

At the same time, stronger multimodal reasoning systems can also be misused. Improvements in visual grounding and image/video understanding may strengthen systems used for surveillance, sensitive visual inference, misinformation analysis, or manipulative image and video editing workflows. The method itself does not introduce a new data source, does not collect private user data, and does not add a new generative capability beyond the underlying models, but it can improve the reliability of existing multimodal policies. Users of released code or checkpoints should therefore follow the safety and usage restrictions of the underlying base models, datasets, and deployment environments.

The evidence-sensitivity analyses in this paper are diagnostic tools rather than safety guarantees. They indicate whether the model's prediction depends on the visual evidence selected by the landing field, but they do not certify factual correctness, fairness, privacy preservation, or robustness under adversarial inputs. Future work should evaluate reward landing under broader safety, fairness, and robustness settings, especially before applying it to high-stakes domains.

\section{Use of Large Language Models}
\label{app:llm_usage}

We used generative AI tools to assist in implementing the methods, including code implementation and experiment execution during the later stages of the project. We did not use generative AI tools to generate synthetic datasets; develop theoretical models or conceptual frameworks; formulate mathematical claims; provide critical ingredients for proving mathematical claims; write proofs; propose or refine hypotheses; design or provide feedback on research methodology or experiments; translate text; clean or reformat datasets; support qualitative or thematic data analysis; or interpret results. The experiments were designed by the authors, who take full responsibility for the final content of this work, including its text, code, and experimental results.

\paragraph{Implementation and experiment execution.} AI coding assistants helped implement components specified by the authors and run training and evaluation jobs. The authors reviewed and tested all AI-assisted code before using it in reported experiments, and checked the configurations, logs, and outputs of every run behind a reported number.

\paragraph{Writing and \LaTeX{} preparation.} AI tools helped organize the \LaTeX{} source and helped check and format references; the authors checked every changed reference against its original source. We also used AI tools to polish the grammar, wording, and flow of text written by the authors; no part of the paper was drafted by AI, and the authors verified and edited all polished text. An example editing prompt is:

\begin{quote}
\textit{Please check the grammar, spelling, word choice, and sentence flow of the following LaTeX text. Keep all equations, labels, references, and macros unchanged, and avoid changing the scientific content.}
\end{quote}

\end{document}